\documentclass{article}

\usepackage{arxiv}

\usepackage[utf8]{inputenc} % allow utf-8 input
\usepackage[T1]{fontenc}    % use 8-bit T1 fonts
\usepackage{hyperref}       % hyperlinks
\usepackage{url}            % simple URL typesetting
\usepackage{booktabs}       % professional-quality tables
\usepackage{amsfonts}       % blackboard math symbols
\usepackage{nicefrac}       % compact symbols for 1/2, etc.
\usepackage{microtype}      % microtypography
\usepackage{lipsum}
\usepackage{natbib,graphicx,graphics,verbatim,xcolor}

\title{Lift Up and Act! \\Classifier Performance in Resource-Constrained Applications}

\author{
  Galit Shmueli%\thanks{Use footnote for providing further
    \\
  Institute of Service Science\\
  National Tsing Hua University\\
  Hsinchu, 30013,  Taiwan\\
  \texttt{galit.shmueli@iss.nthu.edu.tw} \\
}

\begin{document}
\maketitle

\begin{abstract}
Classification tasks are common across many fields and applications where the decision maker's action is limited by resource constraints. In direct marketing only a subset of customers is contacted; scarce human resources limit the number of interviews to the most promising job candidates; limited donated organs are prioritized to those with best fit. In such scenarios, performance measures such as the classification matrix, ROC analysis, and even ranking metrics such as AUC, measure outcomes different from the action of interest. At the same time, gains and lift that do measure the relevant outcome are rarely used by machine learners. In this paper we define resource-constrained classifier performance as a task distinguished from classification and ranking. We explain how gains and lift can lead to different algorithm choices and discuss the effect of class distribution.
\end{abstract}

% keywords can be removed

\section{Introduction: Action-based classifier goals}\label{sec-intro}
Applying machine learning requires defining a precise problem statement in order to train, evaluate, and compare algorithms. The difficulty arises when trying to translate a  domain objective into a machine learning objective. This translation requires not only the operationalization of concepts (e.g., how to measure user satisfaction) but also the choice of performance measures and approach. In this paper we focus on performance evaluation in supervised learning, and specifically for classifiers.

Despite the dominance of several performance measures for comparing and evaluating classifiers, prominent machine learning researchers have emphasized the importance of considering the goal of using classification in order to determine which measures are appropriate and useful and which are not. \citet{hand2009measuring} suggests "Ideally, of course, one would choose a measure which properly reflected one’s aims. Indeed, if the aims have been precisely specified, choosing a measure which does not reflect them could lead to incorrect conclusions, as the different measures need not lead to the same rank-order of performance of classifiers." Similarly, \citet[][p. 84]{japkowicz2011evaluating} emphasize the importance of using domain-relevant performance measures: "The algorithm’s performance needs to be measured on one or a few of the most relevant aspects of the domain, and the performance measures considered should be the ones that focus on these aspects." \citet{flach2019performance} comments "it would be highly desirable, in my opinion, for machine learning experimenters to be more explicit about the objective\ldots and to justify the reported evaluation measure from that perspective." This paper attempts to improve the alignment of classification goal with performance measures by considering the decision maker's \emph{type of action} based on the classifier's predictions.

\subsection{Action-based performance}
Classifiers are used for a variety of goals in different fields and applications. Rather than consider classification performance in isolation, if we consider the \emph{action}  generated by the classification of a new set of records, we can better characterize the underlying goal. We distinguish between three types of goals where classifiers are typically used:
\begin{enumerate}
    \item{\bf Personalized action:}
Many scenarios require making a decision \emph{for each record} in the test set based on its predicted class membership. The timing of classifying each record can be different for different records (e.g., a spam filter is triggered whenever an email comes in). The decision can occur immediately after the predicted score is generated. For example, a financial institution makes (possibly immediate) decisions on whether to grant a loan to each individual applicant, based on the (predicted) risk score/level; The judicial system arrives at a verdict for each sentenced person.  

\item{\bf Batch action:} Some scenarios call for creating a priority list, so that decisions about records can be made in a sequential order that optimizes some criterion. For example, ranking a list of patients arriving at the emergency department allows the medical staff to prioritize care to those in most urgent need, yet eventually all patients will be attended to. Ranking a list of complaints at a power company's call center can help prioritize attending to the most urgent or damaging issues, but eventually all issues will be attended to.

\item{\bf Resource-constrained batch action:}
A third type of action is a resource-constrained action, where the decision maker will only act on \emph{some} of the predicted records. Examples include ranking search query results from most relevant to least relevant to decide which to display on the first results page. Ranking job candidates from most suitable to least suitable in order to allow the hiring committee to most efficiently find the subset of most suitable candidates to shortlist and interview. Ranking college admissions to select those most likely to graduate, given a limited admission capacity. In this scenario, the decision maker has limited time, capacity, attention, budget, or any other constrained resource which does not allow applying the action to all predicted records. In such cases the decision maker's action will only be applied to a subset of the predicted records.
\end{enumerate}

These different goals and utilities of using a classifier translate into different  performance requirements. In personalized action, performance is affected by the \emph{classification of each record} in the test set. In batch action, performance is affected by the \emph{ranking of each record} in the test set. In resource-constrained batch action, performance is affected only by the ranking of the \emph{top ranked records} in the test set, where "top" is determined by the resource limitation.

How do these differences in performance translate into performance measures? In the machine learning literature, the first case is translated into "classification performance" and uses the confusion matrix and various summaries derived from it (e.g., accuracy, recall/precision, sensitivity/specificity, F-measure) and ROC analysis. The second case is translated into "ranking performance", with the most common and relevant measures being the Predicted Positive (PPOS) Rate and Area Under the ROC Curve (AUC) \citep[][p. 348]{flach2012machine}.

Even within classification performance itself,  different measures capture different information and make different assumptions. \citet{provost1998case} criticize the overuse of accuracy, as it assumes equal misclassification costs and a known class distribution in the test set, which is uncommon in reality. Instead, they advocate using ROC curve analysis, which is also useful for combining results from different datasets with different class distributions \citep{flach2012machine}. 

At the same time, \citet{hand2009measuring} criticized the use of AUC for comparing classifiers when their ROC curves intersect, in which case "the AUC uses different misclassification cost distributions for different classifiers."

In a recent paper, \citet{flach2019performance} described the difference between evaluating classification and ranking performance, making the point that different performance measures "assume a different use case: accuracy assumes that, within each class, the difference in cost\ldots is the same, while F-score assumes additionally that true negatives do not add value; both assume that the class distribution in the test set is meaningful. Furthermore, these two measures assume that the classifier has a fixed operating point, whereas AUC\dots deals with ranking performance rather than classification performance." 
\citet{flach2019performance} describes the conflation of classification and ranking measures among machine learners resulting in the reporting of `everything and the kitchen sink'. 

%%%%%%%%%%%%%%%%
\subsection{Common in practice, rare in the literature}
Compared to other classifier performance measures, gains and lift charts, which are designed for resource-constrained classification, rarely appear in the data mining and machine learning literature. In 2009, \citet{burez2009handling} commented that "Apart from their primarily presentational purpose lift charts have not been studied extensively." 
Ten years later, \citet{Wu2019Clarifyin} surveyed papers from the top data mining conferences and journals (with the highest h5-index according to Google Scholar)\footnote{\url{https://scholar.google.com/citations?view_op=top_venues&hl=en&vq=eng_datamininganalysis}} and found a very small number of papers using gains and/or lift chart or analysis\footnote{The search included papers between 2000-5/2019 using keywords "lift chart" OR "gains chart" OR "lift curve" OR "cumulative gains".}, compared to an abundance of papers using other measures of classifier performance\footnote{Keywords for searching for common classifier measures were "ROC" OR "sensitivity" OR "specificity" OR "recall". "Accuracy" was excluded because the term is very broad. Hence the counts of common classifier measures is an under-estimate of non-lift/non-gains measures.}. Table \ref{tab:litsurvey}  \citep[reproduced from][]{Wu2019Clarifyin} shows the results. Given the many resource-constrained ranking applications in practice, the rarity of gains and lift measures in the data mining and machine learning literature is surprising and also alarming.
\begin{table}[h]
    \caption{Number of papers reporting classifier performance measures in top data mining conference proceedings and journals (top h5-index on Google Scholar category "Data Mining \& Analysis" as of 5/2019) - reproduced from \citet{Wu2019Clarifyin}}
    \centering
    \begin{tabular}{l|rr}
    Journal/Conference   & Gains/Lift  &  Other classifier measures \\ \midrule
    ACM Digital Library (SIGKDD, WSDM, TIST, and others) & 98 & 57,310 \\
    IEEE Xplore Digital Library (TKDE, ICDM, and others) & 21 & 99,195 \\
    Knowledge \& Information Systems & 32 & 938 \\
    %International Conference on Data Mining & 41 & 5,840 \\
    Data Mining and Knowledge Discovery & 24 & 445 \\ \midrule 
    \end{tabular}
    \label{tab:litsurvey}
\end{table}

The goal of this paper is threefold: to clarify the distinct nature of constrained-resources batch action classification and contrast it with classification and ranking; to highlight the glaring rarity of performance evaluation for constrained-resources batch action classifiers in machine learning literature (and contests) despite its popularity in many practical applications; and to re-ignite the usefulness of gains and lift charts and analysis in such scenarios.

%%%%%%%%%%%%%%%%%%%%%%
\section{Gains and lift: Notation and computation}% charts and analysis (in a nutshell)}
\label{sec-notation}
The purpose of this section is to describe gains and lift computation, charts, and analysis by introducing notation that helps clarify how they differ from other classifier evaluation measures\footnote{We use both standard machine learning notation and typical statistical notation}. We follow common confusion matrix notation TP, FP, TN, FN to denote counts in each cell: \\
\begin{center}
\begin{tabular}{l|cc}
     & Predicted = Positive & Predicted = Negative \\ \midrule
Actual = Positive  & TP & FN \\
Actual = Negative  & FP & TN \\ 
\end{tabular}
\end{center}

Gains and lift charts have been used mainly in data mining based marketing
campaigns, where direct marketing is a common resource-constrained action, in order "to estimate the payoff of applying modeling to the problem of predicting behavior of some target
population" \citep{piatetsky1999estimating}. Classifiers are typically used to identify, among a set of customers, those who are most likely to perform some action, such as respond to an offer or default on a loan. "Most likely" is often constrained by a restricted number of offers to be made or by some other constraint on the targeted subset of customers. In any case, the goal is to minimize the marketer's effort (and/or the customers' burden) by targeting only a subset of the test set while maximizing some financial (or other) gain. 

Consider a test set (or some population to be scored) with $N$ records. Denote the true class of record $i$ by $y_i$, where $y_i \in \{0,1\}$. A classifier produces for each record $i$ a predicted probability\footnote{For computing gains and lift, the probabilities do not need to be calibrated and any score that preserves the ordering of the records is sufficient.} of belonging to the positive (1) class $\hat{p}_i = \hat{P}(Y_i=1)$. 

\begin{itemize}
    \item Sort the probabilities in descending order $\hat{p}_{(N)},\hat{p}_{(N-1)},\ldots ,\hat{p}_{(1)}$ where $\hat{p}_{(j)}$ is the $j$th largest probability %($\hat{p}_{(N)} = \max (\hat{p}_1,\ldots,\hat{p}_N)$ and $\hat{p}_{(1)} = \min (\hat{p}_1,\ldots,\hat{p}_N)$)
    \item Transform the probabilities into ranks $R(\hat{p}_{(i)}) = N-i+1$, so that rank $j$ corresponds to the record with the $j$th highest probability (e.g., the top ranking record has $R(\hat{p}_{(N)})=1$).
    %\item Denote the true class of record $i$ by $y_i$.
\end{itemize}

\subsection{Cumulative Gains}\label{subsec-cumgains}
The \emph{Cumulative Gains} for a given number of records $n$ is the total number of true positive records among the top-$n$ ranked records: 
\begin{equation}
    \mbox{\it CumGains(n)} = \sum_{j=1}^n y_{(j)} = TP_n.
\end{equation}
Note that we use $n$ as an argument of \textit{CumGains(n)} and as a subscript in $TP_n$ to denote that these quantities are based only on the top-$n$ records.

\subsubsection{Cumulative gains as a sequence of confusion matrices}
Cumulative gains can also be computed using a sequence of confusion matrices, where the cutoff is sequentially set to each of the (unique) $\hat{p}_{(j)}$ values sorted in descending order ($j=N, N-1,\ldots,1$). Equivalently, we can think of a sequence of $n$-confusion matrices ($n=1,2,\ldots,N$) of the form:
\begin{center}
\begin{tabular}{l|cc|c}
     & Predicted = Positive & Predicted = Negative \\ \midrule %\hline
Actual = Positive  & $TP_n$ & $FN_n$=0 & {\it CumGains(n)}=$TP_n$ \\
Actual = Negative  & $FP_n$ & $TN_n$=0 & $FP_n$\\ \midrule %\hline
                    & $TP_n+FP_n$ & 0 & $n=TP_n+FP_n$\\
\end{tabular}
\end{center}
 The $n$-confusion matrix formulation highlights that we are computing the cumulative true positives ($TP_n$) in the top-$n$ records ($\sum_{j=1}^n y_{(j)}$) as a function of the cumulative predicted positives in the top-$n$ records ($n=TP_n+FP_n$). Note that for each $n$, we assume all those $n$ records are predicted as positives and hence the $n$-confusion matrix has zeroes in column Predicted=Negative ($FN_n=TN_n=0$).

\subsubsection{Cumulative gains as a proportion/percentage of total gains}
Cumulative gains are often reported as a proportion or percentage of the total possible gains ($TP+FN$):
\begin{equation}
\label{eq:propgains}
    \mbox{\it p-CumGains(n/N)} = \frac{CumGains(n)}{TP+FN} = \frac{\sum_{j=1}^n y_{(j)}}{\sum_{i=1}^n y_i} = \frac{TP_n}{TP+FN}
\end{equation}
At $n=N$ we get \emph{p-CumGains(1)} equal to \emph{sensitivity} of the classifier. However, typically in constrained-resources applications we are not interested in such large values of $n$.

%%% Chart
\subsubsection{Cumulative gains chart}
\begin{figure}[h]
    \centering
    \includegraphics[width=0.5\textwidth]{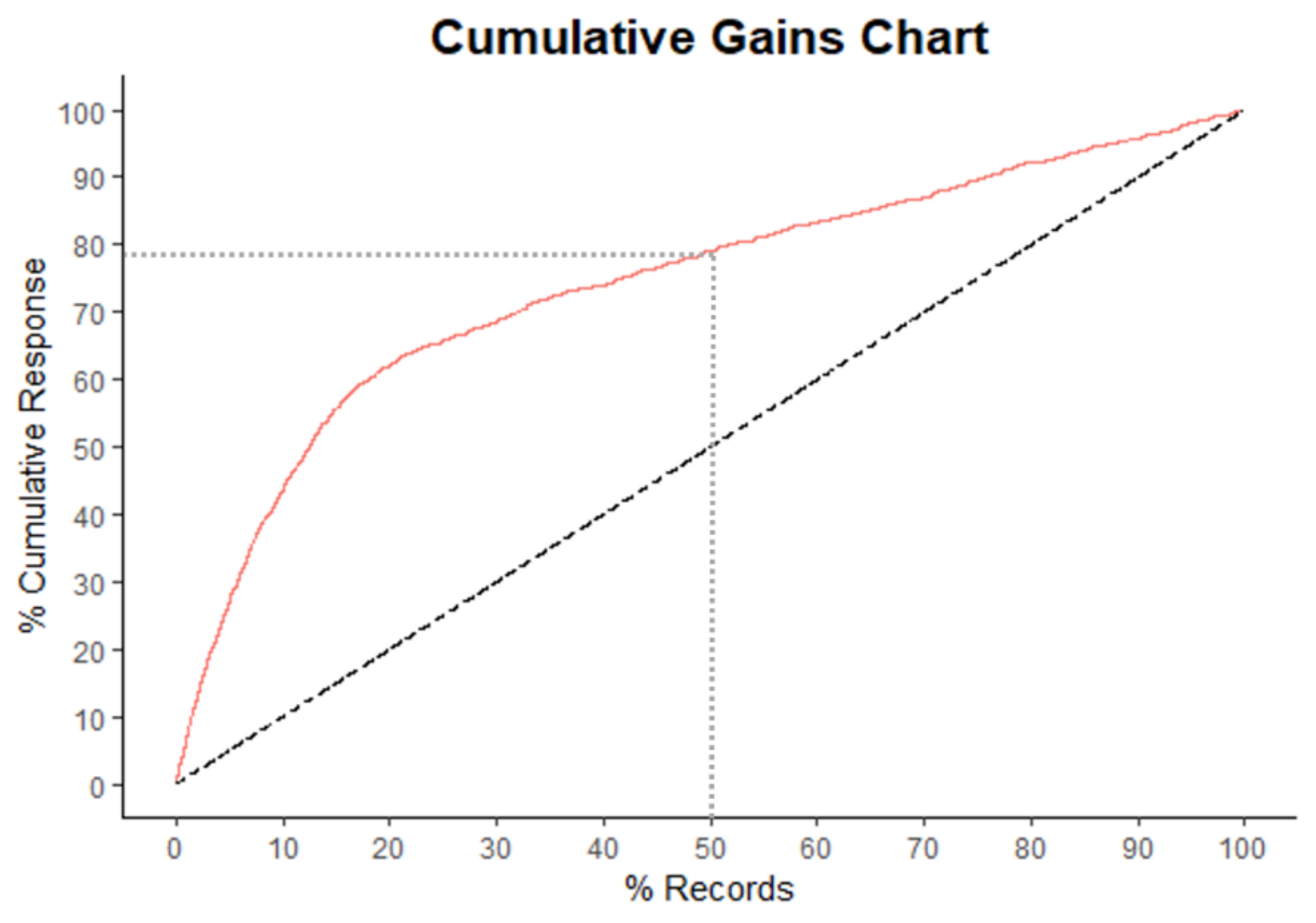} \vspace{0.1in}
    \includegraphics[width=0.5\textwidth]{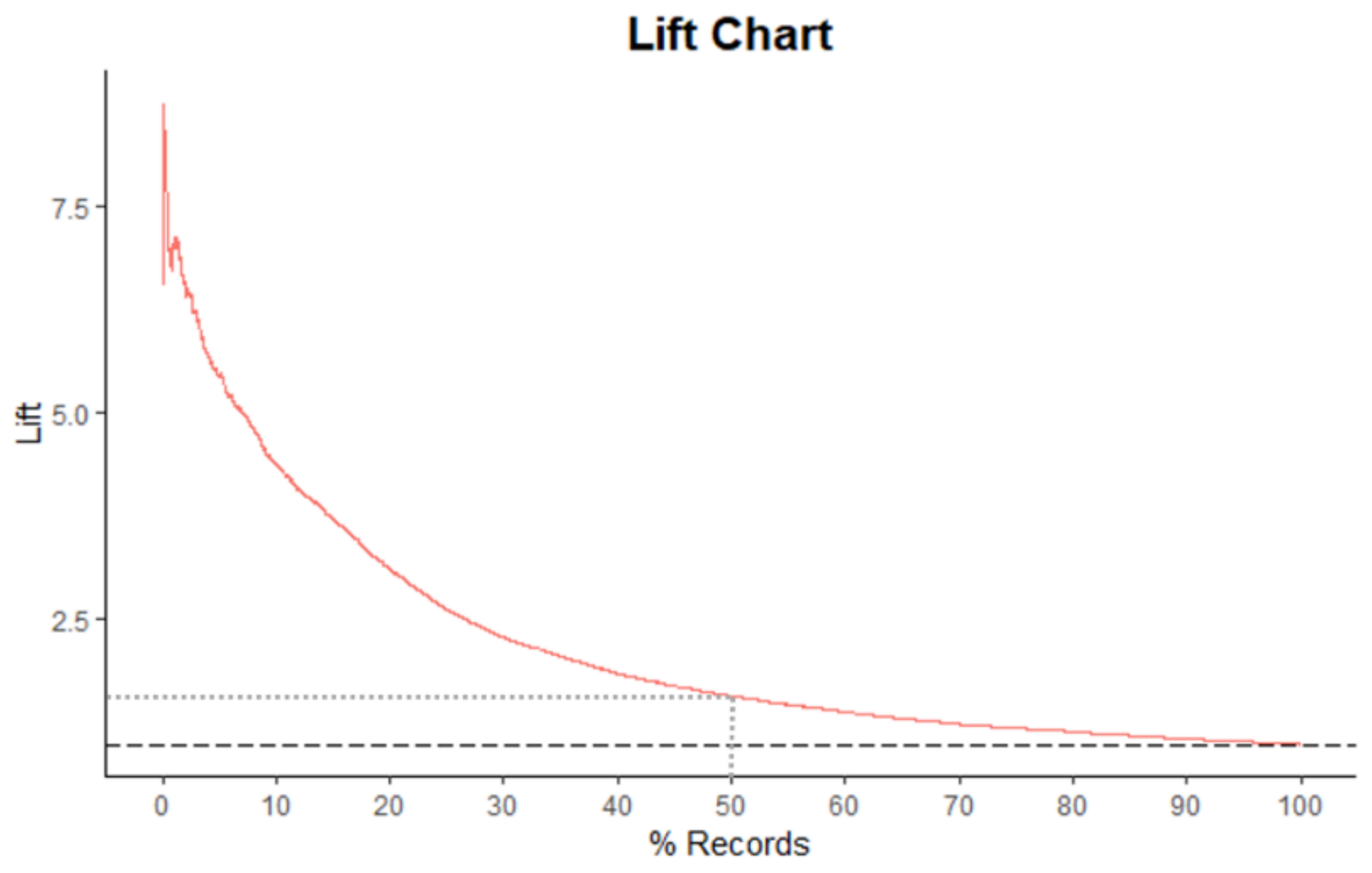} \vspace{0.1in}
    \includegraphics[width=0.5\textwidth]{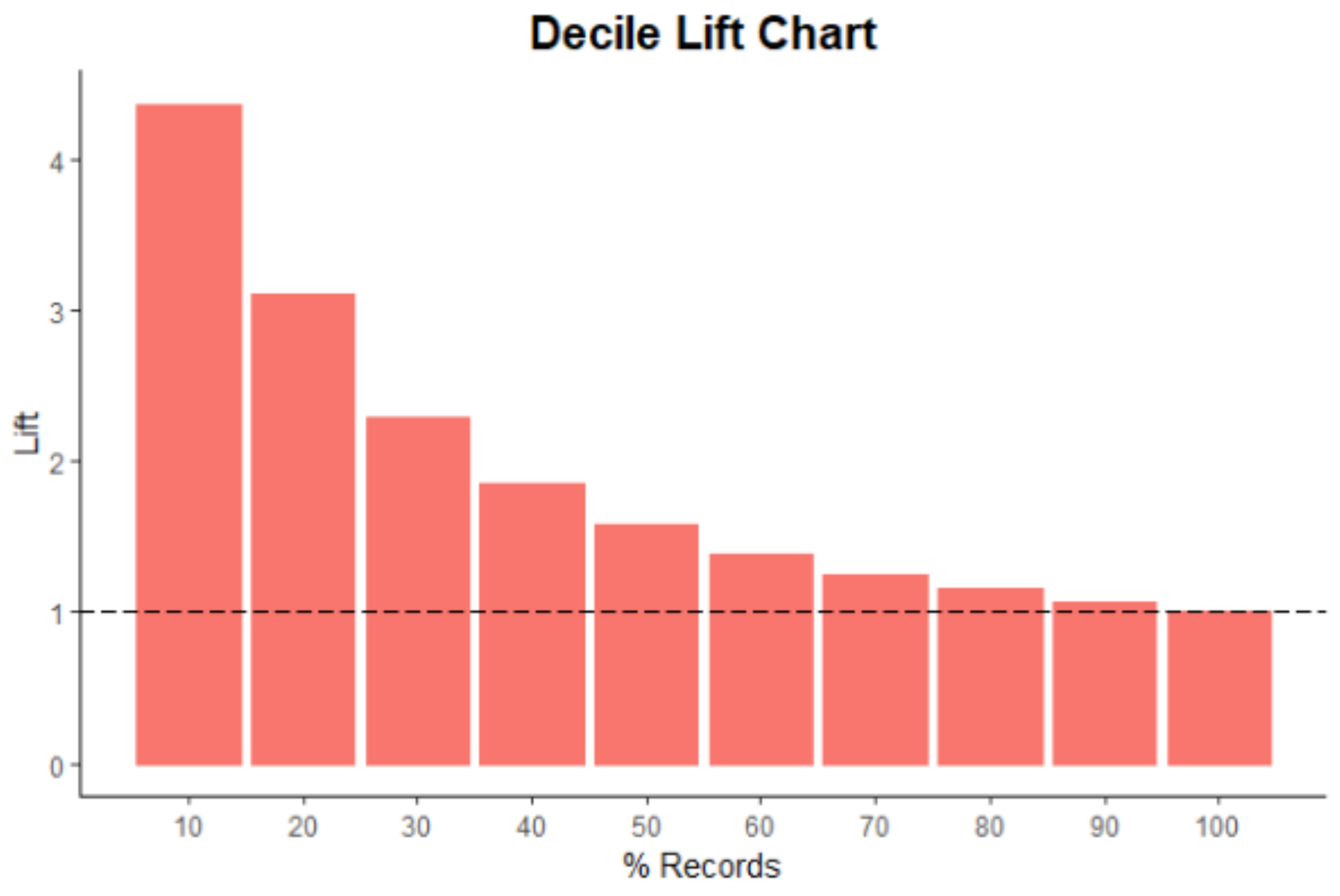}
    \caption{Cumulative gains chart (top),  lift chart (middle), and decile-lift chart (bottom). The dashed line represents random targeting each of the $n$ records with probability $\frac{1}{N}\sum_{i=1}^N y_i$.}
    \label{fig:gainschart}
\end{figure}

\emph{Cumulative gains charts} are plotted in one of two forms:
\begin{enumerate}
    \item Cumulative gains (\emph{CumGains(n)}) are plotted as a function of the targeted number of records ($n$). % in the test set
    \item The proportion/percent cumulative gains (\emph{p-CumGains(n/N)}) is plotted as a function of the proportion ($n/N$) of targeted records.
\end{enumerate} 
 The top panel in Figure \ref{fig:gainschart} shows an example of a percentage cumulative gains chart. We see, for example, that targeting 50\% of the test set records (on the x-axis) is expected to yield nearly 80\% of the total positives (on the y-axis). The diagonal line reflects randomly targeting records with probability equal to the positive class proportion in the test set. Note that this probability is based on the entire test set: %. We therefore use subscripts $N$:
\begin{equation}
\frac{1}{N}\sum_{j=1}^N y_i = \frac{TP + FN}{N} 
\end{equation}

\subsubsection{Cumulative gains with costs/benefits}
When misclassification costs/benefits are available, cumulative gains can take them into account by accumulating the net benefit of the top-$n$ records \citep{provost2013data}:
\begin{equation}
\label{eq:profitgains}
    \mbox{\it CumBenefit(n)} = \sum_{j=1}^n y_{(j)} \cdot q_{TP} + \left(n-\sum_{j=1}^n y_{(j)}\right) \cdot q_{FP}  = TP_n \cdot q_{TP} + FP_n \cdot q_{FP}
\end{equation}
where $q_{TP}$ is the expected net benefit from each TP record and $q_{FP}$ is the expected net benefit (typically negative) from each FP record.

\clearpage
\subsection{Lift}\label{subsec:lift}
\emph{Lift}\footnote{We note a serious confusion between the terms gains, cumulative gains, cumulative response, and lift in the machine learning and related areas. This is reflected in papers and even in software packages that use the terms interchangeably \citep{provost2013data,Wu2019Clarifyin}} is the ratio between the cumulative gains and the corresponding y-value on the diagonal line, which indicates how much better a classifier performs compared to random targeting (in the sense of targeting a proportion $\frac{1}{N}\sum_{i=1}^N y_i$ of the test set at random):
\begin{equation}\label{eq:lift}
    \mbox{\it Lift(n)} = \frac{\mbox{\it CumGains(n)}}{\mbox{\it RandomTargeting(n)}} = \frac{\sum_{j=1}^n y_{(j)}}{\frac{n}{N}\sum_{i=1}^N y_i}  = \frac{TP_n}{\frac{n}{N}(TP+FN)}
    = \frac{TP_n / n}{(TP+FN)/N}
\end{equation}
The higher the lift (and above 1), the better the classifier performs compared to random targeting.  
The right-most formulation displays lift as the ratio of the (true) positives rate in the top-$n$ sample to the positives rate in the entire test set.

A \emph{lift chart} is obtained by plotting the lift ratio from  (\ref{eq:lift}) as a function of $n$ ($n=1,2,\ldots, N$) or $n/N$. The middle panel in Figure \ref{fig:gainschart} shows an example of a lift chart that corresponds to the cumulative gains chart in the top panel. We see, for example, that this classifier performs better than random for targeting up to 80\% of the test set. We can also see the lift for a specific targeted segment size, e.g. targeting 50\% of the test set, results in \emph{lift}$\approx1.6$ ($\approx 80/50$ from the cumulative gains chart), that is, 60\% more positives are captured using the classifier's ranking compared to random targeting.

A common variation is plotting a \emph{decile lift chart}, displaying the lift values for each of the 10 deciles \emph{lift(0.1N), lift(0.2N),}\ldots, \emph{lift(N)}. The bottom panel in Figure \ref{fig:gainschart} shows the corresponding decile-lift chart, where we can more easily see that targeting, say, the top decile of the test set results in a lift of 4.5, whereas targeting 50\% of the test set results in a lift of 1.6.

\subsection{Gains and lift charts analysis}
We note that when $n=N$ we get \emph{lift = sensitivity (=recall)}. This would correspond to a scenario with no resource constraint. 
However, importantly, in gains and lift charts, not all areas are equally important. Because the decision maker's action will start from records with the highest $\hat{p}_i$ scores, the left part of the charts is more critical, and performance is expected to deteriorate as we move to the right (thereby acting on an increasingly larger proportion of the test set).

The gains chart can be used for two purposes: if the resource constraint is fixed in advance, then we can evaluate and compare the performance of classifiers when used for targeting the top-$n$ records. It is possible, for example, that one classifier performs best (in terms of cumulative gains) at targeting the top 10 customers, but another classifier outperforms it for the next 100 customers.
A second use is when the resource constraint is more flexible, in which case the gains chart can help determine a useful value $n$ or $n/N$ for a classifier.

\citet{provost2013data} note that the lift curve for each algorithm is based on its own ranking of the records. Hence, when comparing lift curves of multiple algorithms, although different algorithms might have the highest lift for different $n$, one cannot combine them in the simplistic fashion of "use algorithm A for targeting the first $n_1$ records, and then use algorithm B for targeting the next $n_2$ records". Instead, the algorithms can be combined using an ensemble and then the lift chart of the ensemble is evaluated.

\section{Classifier selection based on lift/gains vs. accuracy, ROC, and AUC}
\label{sec:headings}

In the case of an ideal classifier that perfectly classifies each record in the test set, all performance measures -- measuring classification, ranking, and constrained ranking -- would be at their ultimate value. However, in the more realistic scenario of multiple non-perfect classifiers, the three different types of measures can lead to different classifier selections.

In their criticism of using overall accuracy to compare classifiers where the goal is to select the classifier with the lowest cost, \citet{provost1998case} consider two candidate justifications (1) whether a classifier with the highest accuracy might minimize cost, and (2) whether the algorithm producing the highest-accuracy classifier may produce the lowest cost by training it differently.

While batch action and resource-constrained action are clearly different in terms of their real-world action and implications, the question is whether model selection (choosing a classifier among a set) using classification performance measures (accuracy, ROC curve) or ranking performance measures (AUC) would lead to the same choice as a resource-constraint measure (gains, lift).

The difference between unconstrained and constrained ranking is strongly pronounced when costs/benefits are considered.
\citet[][pp. 213-214]{provost2013data} show an example of profit curves for three classifiers (equivalent to a gains chart that integrates costs), where one classifier is optimal in terms of maximizing profit when there is no resource constraint, while a different classifier is optimal when a budget constraint is imposed. This simple visual example highlights the issue of the constrained optimization space and how it differs from the unconstrained space.
\begin{figure}[h]
    \centering
    \includegraphics[width=0.5\textwidth]{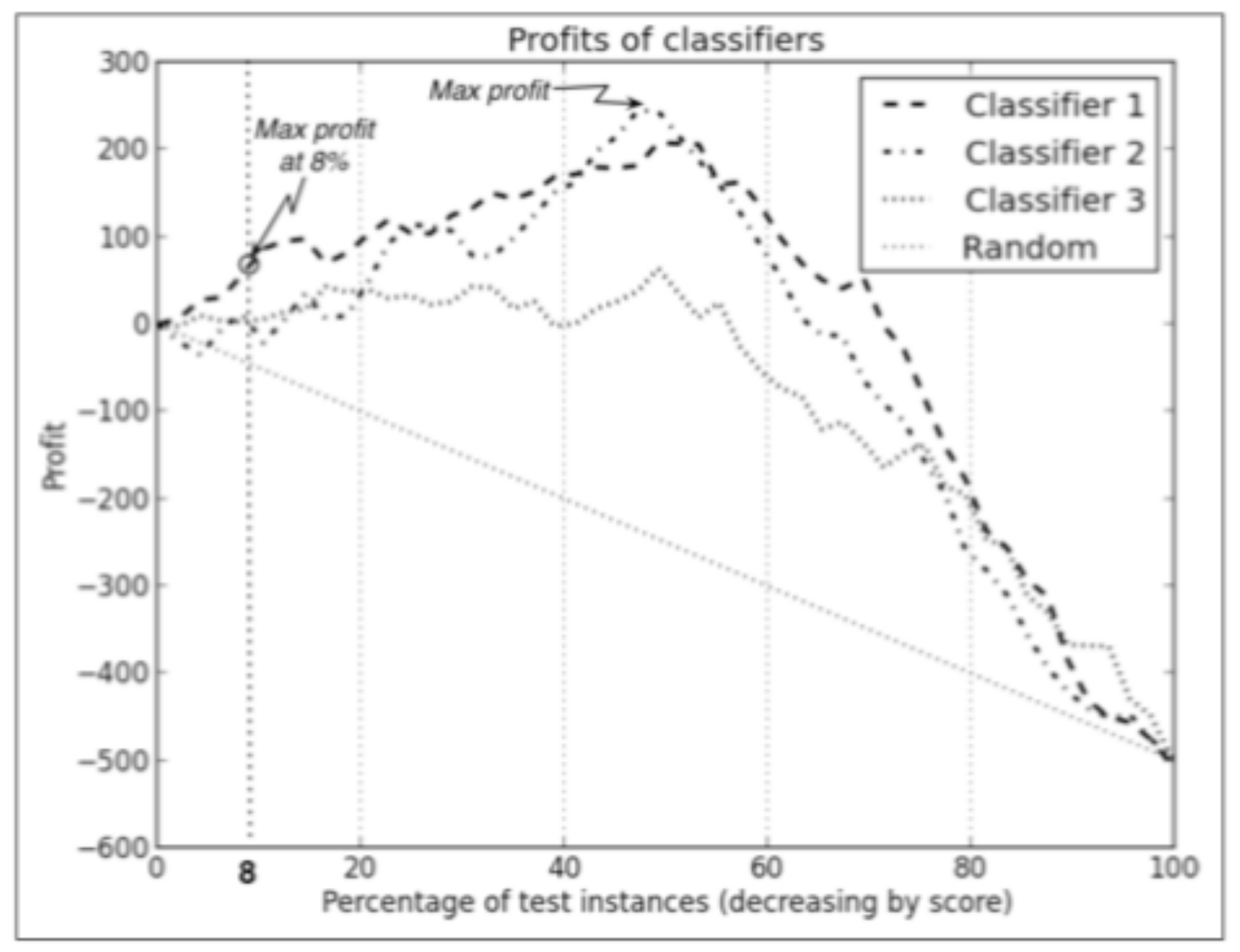}
    \caption{[From \citet{provost2013data}] Profit curves for 3 classifiers: Classifier 3 is best for unconstrained profit maximization, while Classifier 1 is best if budget is constrained to targeting 8\% of the records.}
    \label{fig:my_label}
\end{figure}
%\newpage

We therefore ask four questions for classifier selection:
\begin{enumerate}
    \item Can a classifier with lower accuracy have higher lift?
    \item Will a classifier with best ROC performance always have highest lift?
    \item Will a classifier with higher AUC always have higher lift? 
    \item Will a classifier with higher lift in the top-$n$ records (e.g. top percentile) always have higher AUC?
\end{enumerate}

\subsection{Lift vs. accuracy}
\emph{Question: Can a classifier with lower accuracy have higher lift?} \\ 
\emph{Answer: Yes.} \\

\citet{piatetsky1999estimating} explain:
\begin{quote}
    "while accuracy measures correct predictions for the whole population, lift measures the increased accuracy for a targeted subset, e.g. the top part of a model-score ranked list. Therefore it is very possible that a modeling method with lower accuracy can result in higher lift at the top of the list."
\end{quote}

%%%%%%%%%%%%%%
\subsection{Lift chart vs. ROC chart}
\emph{Question: Does selecting the classifier with best ROC performance guarantee highest lift?}\\
\emph{Answer: No.}\\

\citet[][Theorem 11]{provost2001robust} shows that for a set of classifiers, the point in ROC space corresponding to the highest lift for some $n/N$ will not necessarily be on the ROC curve of any of the individual classifiers. Unlike maximizing accuracy or minimizing expected cost, where the optimal classifier will be a point on one of the classifiers' ROC curves, the optimal lift point might not be on the ROC curves of any of the existing classifiers\footnote{\citet{provost2001robust} prove this using the notions of \emph{ROC convex hull} (ROCCH) -- the convex hull of the set of points in ROC space, and the \emph{ROCCH-hybrid} -- the set of classifiers that form the ROCCH. The lift maximizing point in ROC space might be on a connecting segment of the ROCCH, and not on any of the actual ROC curves of the classifiers in the comparison set.}.

This theorem implies that when comparing a set of classifiers using ROC, selecting a model by "best ROC" will not necessarily lead to best lift at any specific value $n/N$.

%%%%%%%%%%%%%%%
\subsection{Lift vs. AUC} 
\emph{Question: Will a classifier with higher AUC always have higher lift? Or, does selecting the classifier with highest AUC guarantee highest lift?}\\
\emph{Answer: No.}\\

Although both AUC and gains/lift are affected by the order of the positives and negatives in the vector of actual values (sorted by predicted probability), AUC is affected by the order in the entire vector, whereas \emph{CumGains(n), p-CumGains(n/N)} and \emph{lift(n)} are sensitive only to the order in the first $n$ records (or, $n/N$ proportion of records). Therefore, changing the order of positives and negatives affects AUC and gains/lift differently based on \emph{where} the change is made. 

\begin{figure}[]
    \centering
    \includegraphics[width=3in]{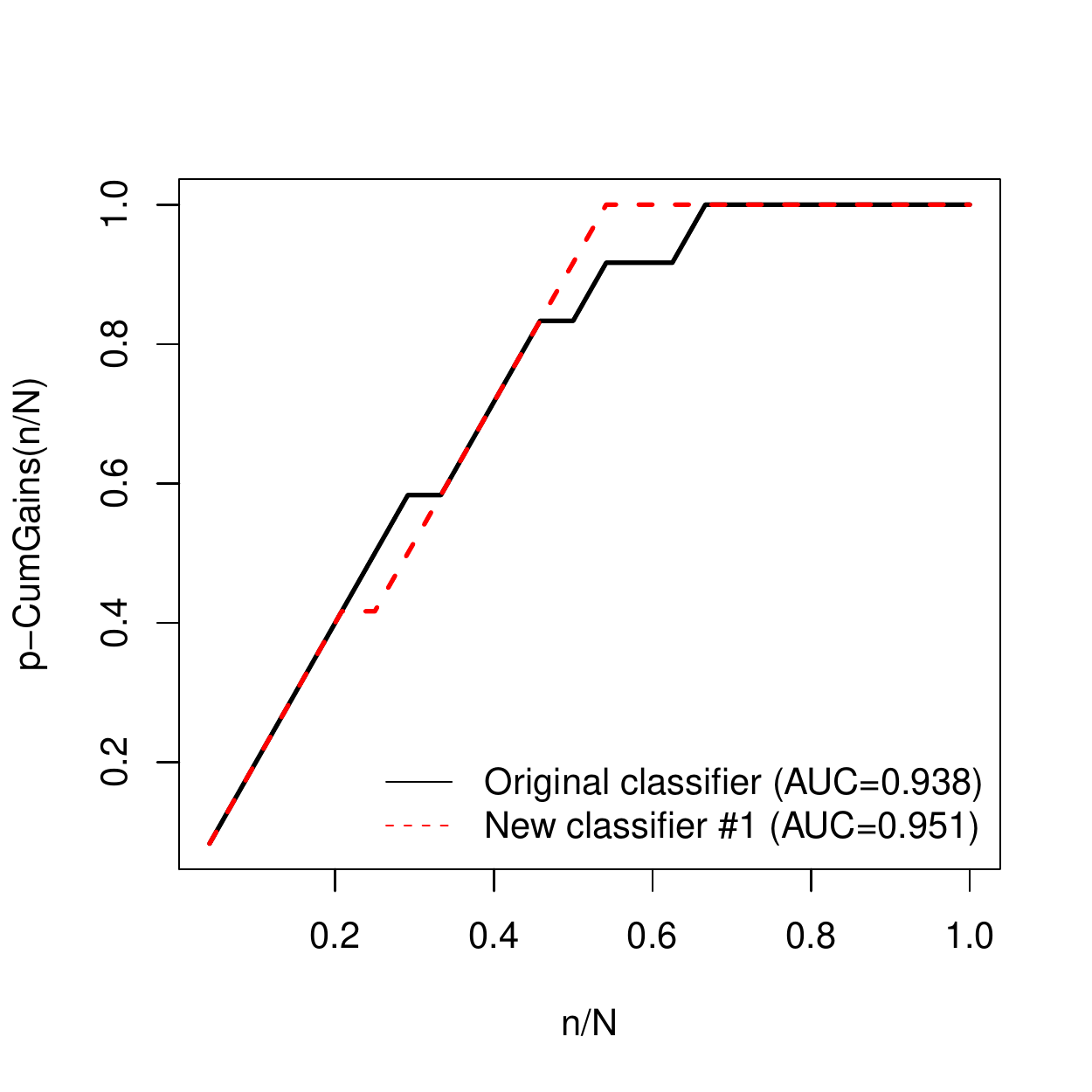}
    \includegraphics[width=3in]{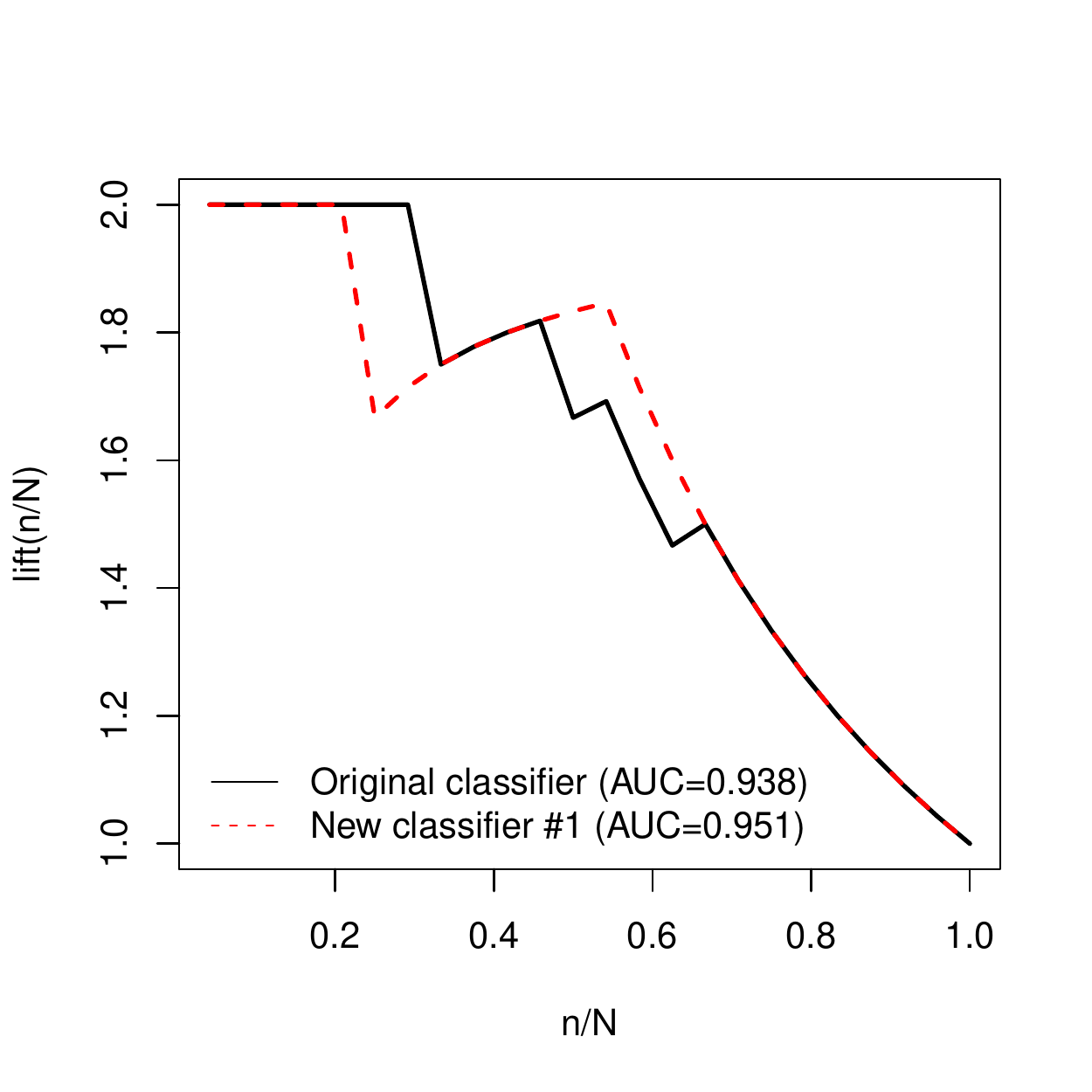}
    \caption{Cumulative gains and lift charts of original classifier (test set from Table \ref{tab:example}) and new classifier} % the same set where $y_{(16)}$ and $y_{(17)}$ are flipped (red; classifier \#1)}%. The curves are identical until $n/N=15/24$.}
    \label{fig:messed}
\end{figure}

Consider the small example of a ranked test set (N=24) shown in Table \ref{tab:example}. For this sample, we get AUC=0.938. The gains and lift charts are shown in Figure \ref{fig:messed} (solid black lines). 
\begin{table}[ht]
\caption{Hypothetical test set, sorted by some classifier from highest score (top) to lowest score (bottom). Bold rows are ranked differently by the new classifier}
\centering
\begin{tabular}{rrrrr}
Rank (\emph{n})  & Actual & \emph{CumGain} & $n/N$     & \emph{p-CumGain} \\ \hline
1  & 1           & 1           & 0.04167 & 0.08333     \\
2  & 1           & 2           & 0.08333 & 0.16667     \\
3  & 1           & 3           & 0.12500 & 0.25000     \\
4  & 1           & 4           & 0.16667 & 0.33333     \\
5  & 1           & 5           & 0.20833 & 0.41667     \\
\bf 6  & \bf 1           & \bf 6           & \bf 0.25000 & \bf 0.50000     \\
7  & 1           & 7           & 0.29167 & 0.58333     \\
\bf 8  & \bf 0           & \bf 7           & \bf 0.33333 & \bf 0.58333     \\
9  & 1           & 8           & 0.37500 & 0.66667     \\
10 & 1           & 9           & 0.41667 & 0.75000     \\
11 & 1           & 10          & 0.45833 & 0.83333     \\
\bf 12 & \bf 0           & \bf 10          & \bf 0.50000 & \bf 0.83333     \\
13 & 1           & 11          & 0.54167 & 0.91667     \\
14 & 0           & 11          & 0.58333 & 0.91667     \\
15 & 0           & 11          & 0.62500 & 0.91667     \\
\bf 16 & \bf 1           & \bf 12          & \bf 0.66667 & \bf 1.00000     \\
17 & 0           & 12          & 0.70833 & 1.00000     \\
18 & 0           & 12          & 0.75000 & 1.00000     \\
19 & 0           & 12          & 0.79167 & 1.00000     \\
20 & 0           & 12          & 0.83333 & 1.00000     \\
21 & 0           & 12          & 0.87500 & 1.00000     \\
22 & 0           & 12          & 0.91667 & 1.00000     \\
23 & 0           & 12          & 0.95833 & 1.00000     \\
24 & 0           & 12          & 1.00000 & 1.00000     \\ \hline
\end{tabular}
\label{tab:example}
\end{table}

%\clearpage
We now create a new classifier by slightly perturbing the order of four rows in the data:
the new classifier switches the actual values in ranks 6 and 8 and also switches the order of actual values in ranks 12 and 16. Compared to the original classifier, this classifier has $y_{(6)}=0, y_{(8)}=1, y_{(12)}=1, y_{(16)}=0$. The new classifier results in AUC=0.951 (higher than the original classifier). Figure \ref{fig:messed} compares the cumulative gains and lift of the new classifier to the original classifier. In terms of lift, it performs worse for $0.2<n/N<0.375$, better for $0.458<n/N<0.667$, and equally well for all other values of $n/N$. As this example illustrates\footnote{More extreme examples can be constructed, especially when performance is less ideal than in this example.}, choosing between classifiers in constrained-resources scenarios should not be guided by AUC, but rather by gains/lift in order to consider the constraint ($n$) and/or required \emph{p-CumGains}. In other words, model selection by AUC risks inferior lift.

\emph{Question: Will a classifier with higher lift in the top-$n$ records (e.g. top percentile) always have higher AUC?}\\
\emph{Answer: No.}\\

\begin{figure}[h]
    \centering
    \includegraphics[width=3in]{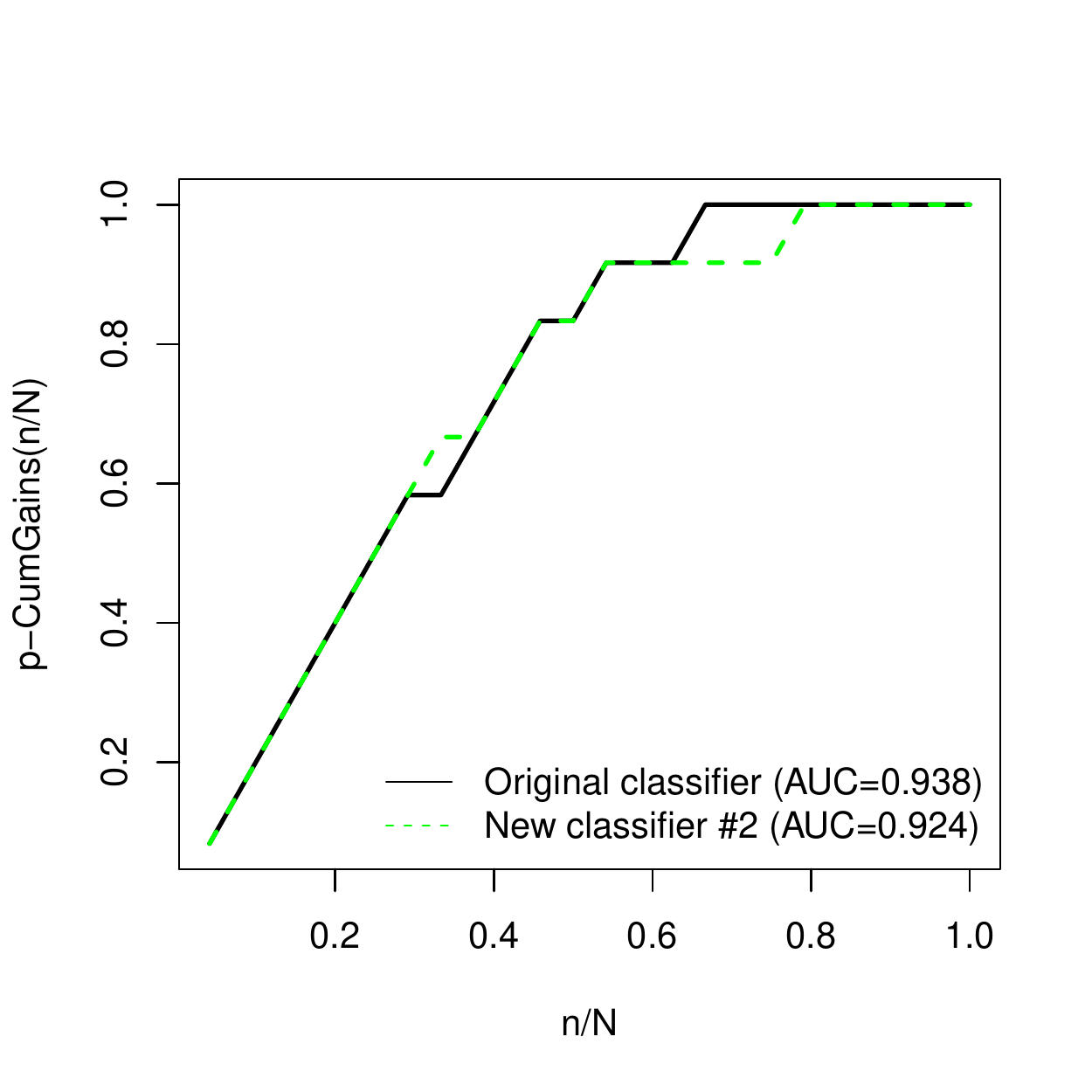}
    \includegraphics[width=3in]{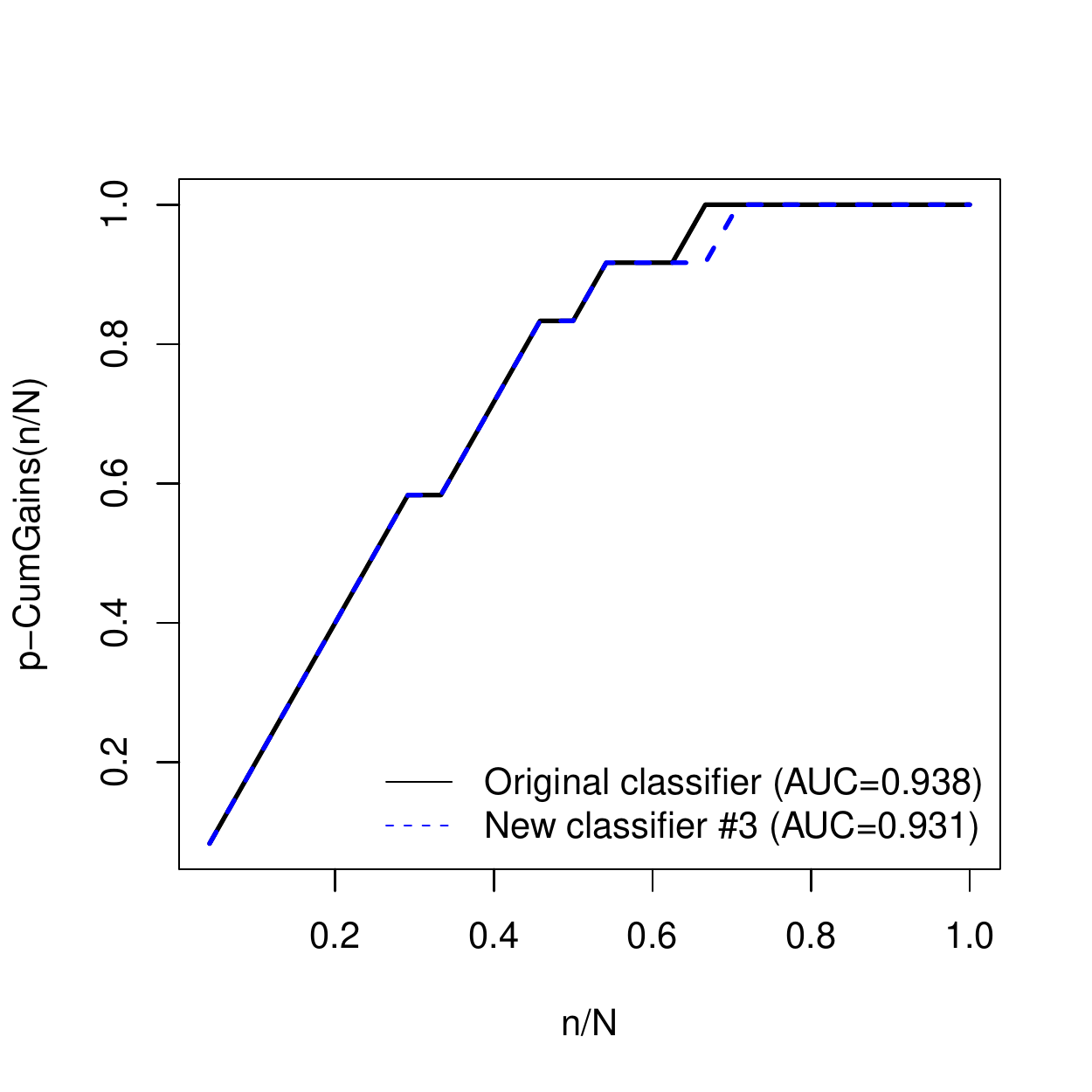}
    \caption{Cumulative gains charts for two classifiers obtained by switching 0,1 values in Table \ref{tab:example}}
    \label{fig:2examples}
\end{figure}

We can construct a new classifier with AUC lower than the original classifier's AUC that results in higher lift for some ranges below $n/N$, and lower lift for larger values of $n/N$. New classifier \#2 in Figure \ref{fig:2examples} (left panel) illustrates such a classifier (this classifier flips the order of ranks 8 with 9 and ranks 16 with 19). Although its lift is better or equal to the original classifier until $n/N=0.667$ its AUC is lower. In such cases too, model selection by AUC risks inferior lift in the relevant area. 

Finally, we might have a classifier with slightly lower AUC but the same lift until very high values of $n/N$ (see new classifier \#3 in Figure \ref{fig:2examples}, right panel). In such cases choosing the highest AUC classifier would not be wrong in terms of gains/lift, but it ignores the equal lift performance in the relevant area (low values of $n/N$), thereby missing the opportunity to select on other criteria such as algorithm simplicity, computation time, ease of implementation, and classifier interpretability.

\citet{martens2016mining} provide an example that illustrates the answers to both the lift-vs-AUC questions. They compared several classifiers applied to a large dataset on bank customers with fine-grained behavioral data, for predicting consumer purchase likelihood of two financial products. They  found that "In terms of lift, the best AUC model (BeSim with the Beta distribution tuned for AUC) does not perform well" and show that the classifier achieving the highest lift at $n/N=0.01$ is different from the classifier with the highest AUC. In fact, the combination of these two classifiers (best AUC and best lift(0.01)) achieved an even better AUC, although in terms of lift it was inferior to the single best-lift(0.01) classifier.

%%%%%%%%%%%%%%%%%%%%%%%

\subsection{Formulas for AUC vs. cumulative gains}
While ranking the records in descending order is required in the two batch actions, the resource-constrained context calls for a performance measure different from AUC. Specifically, AUC is equal to the probability that a random positive record will rank (correctly) higher than a random negative record. AUC is thus proportional to the number of correctly ordered pairs %. Using 
%More formally, AUC can be written in different forms 
\citep{hanley1982meaning, flach2012machine}. Using our earlier notation we can write this interpretation of AUC as:
\begin{equation}
    AUC = \frac{1}{N^+ N^-} 
  %  \sum_{1}^{N^+} \sum_{1}^{N^-} I\left[\hat{p}(x)^+ > \hat{p}(x)^- \right] + \frac{1}{2}I\left[ \hat{p}(x)^+ = \hat{p}(x)^-\right] \\
    %&=&    
    \sum_{i=1}^{N-1} \sum_{j=i+1}^N \left( I\left( y_{(i)} > y_{(j)}\right) - \frac{1}{2} I\left( \hat{p}_{(i)} = \hat{p}_{(j)}\right) \right)
\label{eq:AUCflach}
\end{equation}
where $N^+$ and $N^-$ are the number of positive and negative records, respectively\footnote{The summation counts the number of correctly ordered pairs and discounts ties of positive and negative records which have the same predicted probability.}.   %$\hat{p}(x)^+$ is the probability of a positive record, and $R(x)^+$ is the rank of a positive record (and similar notation applies to the negative records).

AUC is proportional to the Mann-Whitney-Wilcoxon U statistic where $U$ is the sum of positive ranks (also known as the Wilcoxon rank-sum statistic) \citep{hand2009measuring} that sums the ranks of the positive records, where the smallest rank is assigned to the smallest predicted probability (the opposite of our notation in Section \ref{sec-notation}).  We can therefore write AUC\footnote{Ranks of ties are equal to the midpoint between the ranks.}: % as a quantity proportional to the sum of the ranks of the positive records:
\begin{equation}
    AUC = \frac{U}{N^+ N^-} = \frac{1}{N^+ N^-}
    \left( \sum_{j=1}^N y_{(j)} R(\hat{p}_{(N-j+1)})  -\frac{N^+(N^+ + 1)}{2} \right)
\label{eq:AUCwilcoxon}
\end{equation}
Equation (\ref{eq:AUCflach}) shows that we are comparing \emph{all} pairs of positive and negative records. Equation (\ref{eq:AUCwilcoxon}) sums over \emph{all} ranks of the positive records. In both interpretations, the computation involves computing quantities based on the entire set of positive records (and implicitly, the negative records). In contrast, in gains and lift only the top-$n$ ranked records are involved in the computation for every $n$, and except for the right-most part of the chart where $n=N$, gains/lift values are based on only a subset of the records.

This means that reordering positive and negatives at ranks below $n$ will not affect \emph{CumGains(n), p-CumGains(n/N)} and \emph{lift(n)}, but it will affect AUC. Therefore classifiers with different AUC values can have the same (percentage) gains or lift curves for $n/N=0,\ldots c$. This difference reflects the fact that performance of ranks below some threshold are of no interest to the decision maker in resource-constrained tasks, whereas in ranking/discrimination tasks all records matter.

%%%%%%%%%%%%%%%%%%%%%%

\section{Gains, Lift, and Class Distribution}
In the literature, the main criticism of gains and lift charts is their dependence on class distribution (prior distribution) whereas, for example, ROC and AUC are independent of class distribution. To clarify, when discussing dependence on class distribution, the question of interest is how a classifier, which was trained on a certain class distribution, will perform on a test set with a different class distribution\footnote{We note that a different question is the performance of a classifier when \emph{trained} on different class distributions. For example, performance when trained using a random sampling vs. over-sampling. In such a case, even ROC/AUC would be affected.}. This question is relevant when the classifier will be deployed to data with a different (or unknown) class distribution. The underlying assumption is that the misclassification error rates remain constant irrespective of the class distribution.  

We claim that this criticism does not justify not using gains/lift, or replacing them with ROC or AUC, for several reasons:
\begin{enumerate}
    \item Lift and gains charts are appropriate for a goal that is different from classification or ranking: they are aligned with a resource-constrained goal. Aligning the goal and the performance measure is of utmost importance for successful implementation and use in practice.
    \item Lift and gains charts are useful even if the number of positives in the test set is unknown, a case that can be common in practice \citep{vuk2006roc}.\footnote{\citet{vuk2006roc} list the unavailability of the number of potential positives (customers) as a motivation to use gains charts (which they mistakenly call "lift charts") over ROC, because in such cases sensitivity cannot be computed.}
    \item Although there is (currently) no theoretical derivation of gains or lift for a test set with a different response rate, we will show several claims that can be made, as well as illustrate an empirical approach to evaluate such performance. 
\end{enumerate}

\subsection{Empirical evaluation under different class distributions}
We can evaluate the performance of a test set with a different class distribution by taking stratified samples from the test set with the class distribution of interest. To illustrate this, we used the Bank Marketing dataset from UCI Machine Learning Repository\footnote{\url{http://archive.ics.uci.edu/ml/datasets/Bank+Marketing}} where customers of a Portuguese bank received a marketing call with an offer for a bank term deposit; each customer either accepted (subscribed=yes) or declined (subscribed=no) the offer. A typical goal is to build a classifier for targeting other/future customers with this product. The dataset has 45211 records with 11.7\% positives. We randomly partitioned it into training (70\%) and test (30\%). A random forest trained on the training set was used to score the test set. Next, to compare the gains and lift performance of this classifier on a different class distributions, we randomly generated three types of sub-samples, each with size N=5000: (1) 11.7\% positives (random sample from the test set), (2) 5\% positives (stratified sample from the test set), and (3) 20\% positives (stratified sample from the test set). We generated 50 sub-samples of each type and plotted the \emph{p-CumGains} and \emph{lift} in Figure \ref{fig:bank-subsamples}. 

\subsection{Some regularities}
\begin{figure}[h]
    \centering
    \includegraphics[width=3in]{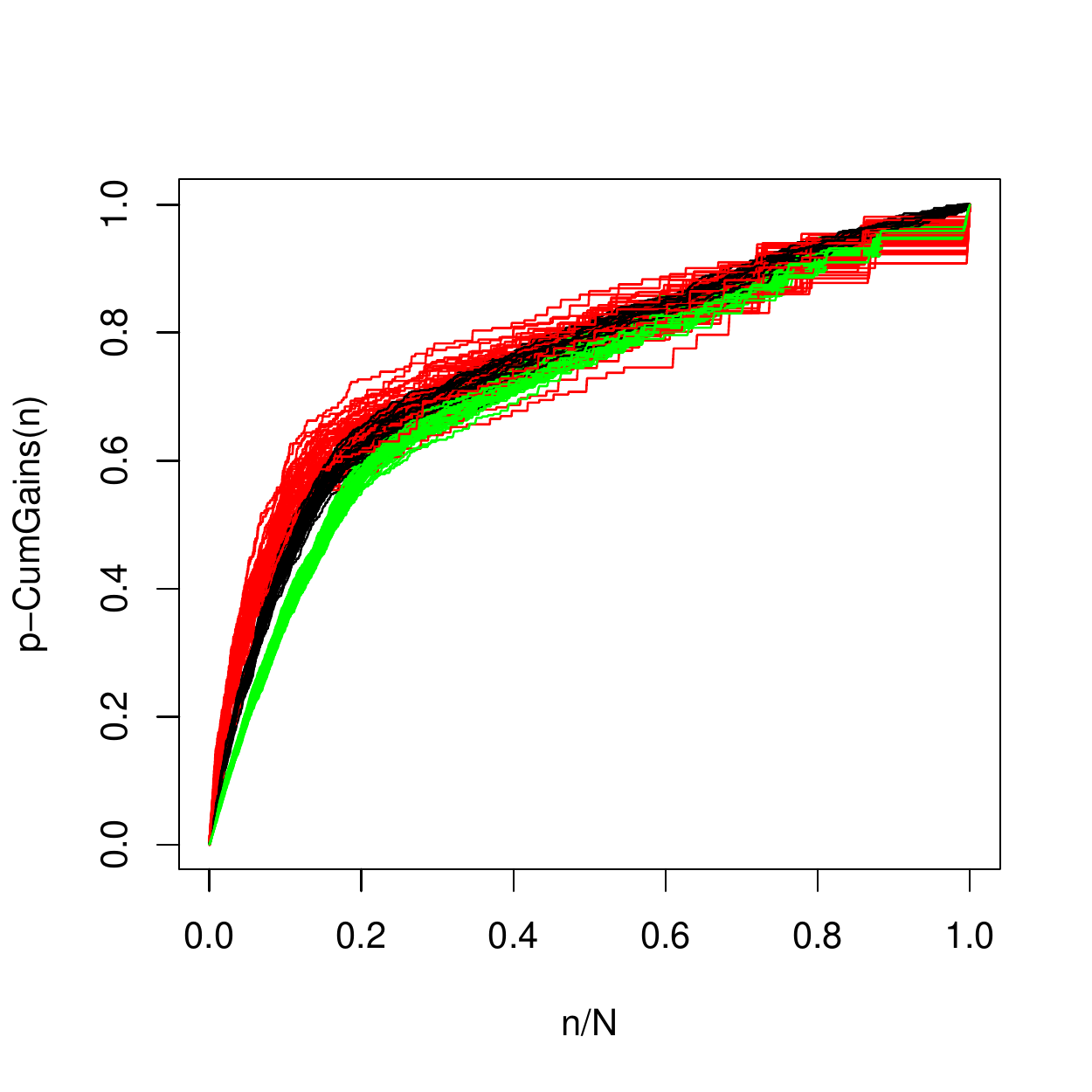}
    \includegraphics[width=3in]{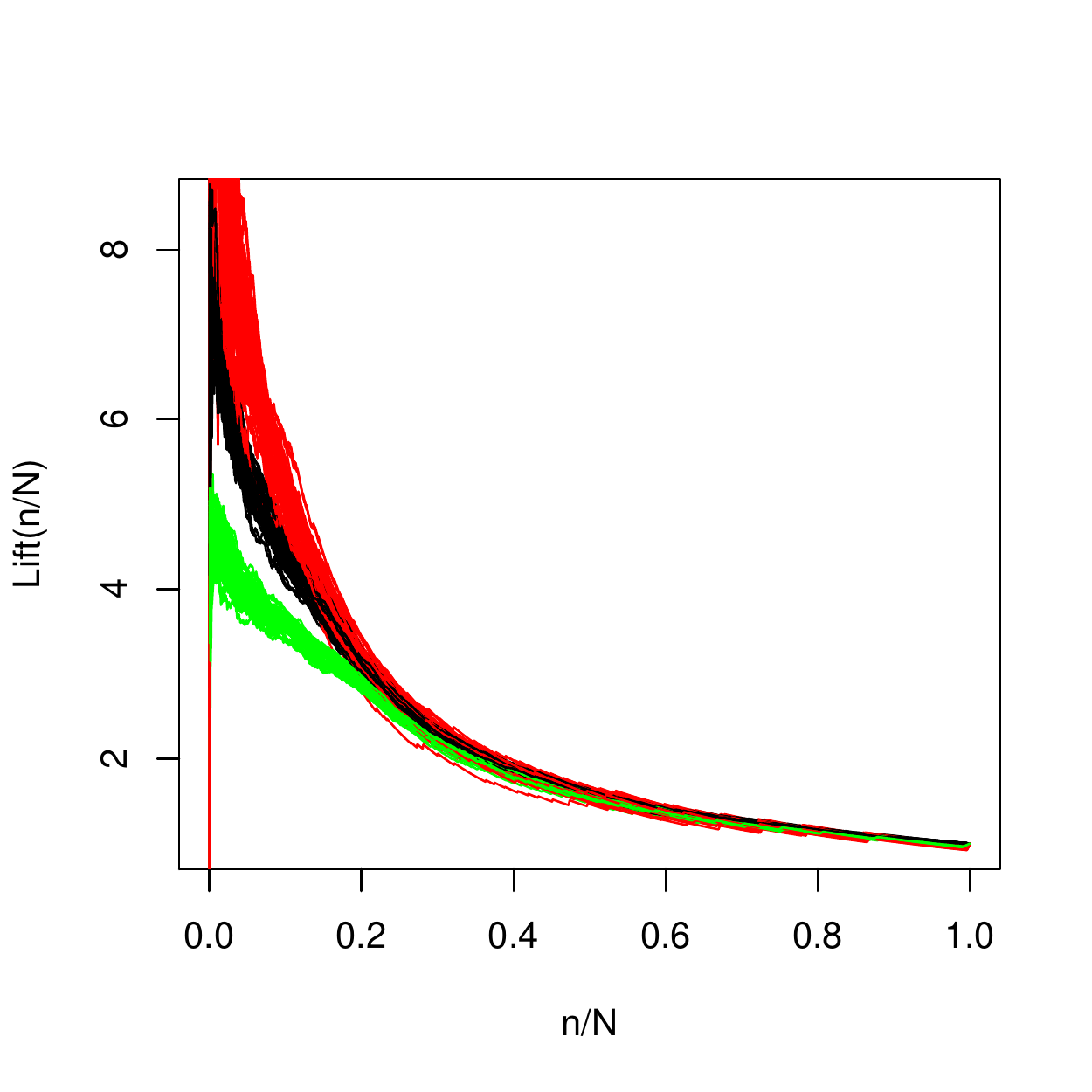}
    \caption{p-CumGains chart (left) and lift chart (right) for 50x3 test sets (each N=5000) sampled from a larger test set (N=12356), with different proportions of positives (black=0.11, red=0.05, green=0.20)}
    \label{fig:bank-subsamples}
\end{figure}

We see from Figure \ref{fig:bank-subsamples} that at lower targeting values (left part of the chart), gains and lift are affected by class distribution: deploying the classifier in a lower percent positives scenario (rarer positives) is associated with better performance, while deployment to populations with more balanced class distribution is expected to have lower gains/lift performance. As a larger proportion is targeted these differences get smaller and eventually disappear. This approach allows evaluating gains and lift performance on different percent positives of interest, for the purpose of testing sensitivity as well as for setting expectations for the decision maker under different class distribution scenarios.

In fact, we can say more about the effect of class distribution: \emph{scoring a test set with a lower rate of positives will lead to higher lift performance for small $n<N$ for any classifier that is better than random}. We can see this by considering that the AUC is insensitive to class distribution, that is, a classifier applied to two test sets with positives rates $p$ and $p'<p$ will have the same AUC. Because the AUC is the probability that a random positive record is ranked higher than a random negative record \citep{hanley1982meaning}, it implies that the ranks of the positive records in both test sets will be similar, especially in the top positions. We can see this from the formulation of AUC as the average number of pairs where a positive record is ranked higher than a negative record (equation \ref{eq:AUCflach}).

Consider an $N$-vector of ranked labels with positives rate $p$. To get an $N$-vector with $p'<p$ we can think of the same vector with $N(1-p'/p)$ of the positive records replaced by negative records. The new  negative records, if scored by a better-than-random classifier, would tend (on average) to have ranks lower than the removed positives. More accurately, the ranks of a "replacement" negative must be lower than the next-lower-ranked positive record in order to achieve the same AUC as the $p$ test set. %Achieving the same AUC requires that an equal probability of a positive record exceeding a negative record in both test sets. 

The implication of the above statement is twofold: first, that gains and lift performance for low $n/N$ values (where the positives are ideally concentrated) will be better for lower $p'$. Second, the better the lift of a classifier for $p$, the better the improvement in gains and lift (for low $n/N$) for a lower  $p'$, thereby providing another model selection criterion.

%%%%%%%%%%%%%%%%%%%

\section{Conclusions}
We distinguish resource-constraint performance measures (gains and lift) from classification and ranking, based on the decision maker's action. Using the decision maker's action for determining appropriate performance measures better aligns the domain and data mining goals and their utility. In this sense we extend and diverge from current guidelines for choosing between different metrics. We do not simply see gains and lift charts as "easier to read" for non data scientists as compared with ROC curves, but rather as a measure adequate for a different goal. \citet{provost2013data} advocate using lift and cumulative gains charts when cost information is unavailable yet the class distribution is known and fixed; when the class distribution is unknown or expected to change, they advocate ROC curves. Although class distribution affects gains and lift calculations, they can be modified directly using different scenarios (e.g., by creating sub-samples using stratified sampling). Therefore, rather than resorting to using ROC curves, we advocate using gains and lift charts for resource-constrained tasks even when the class distribution is unknown or expected to change. In that sense, we extend the distinction between \emph{classification} and \emph{ranking} by \citet{flach2012machine} to include a third category: \emph{resource-constrained ranking}. Whereas classification warrants the use of the confusion matrix, its summaries and ROC curves for evaluating performance, ranking can be measured by AUC, while for resource-constrained ranking gains and lift are the most appropriate.

Future useful directions include possibilities of tailoring the training of classifiers to optimize lift and comparing such an approach to ordinary optimization of error or ranking. \citet{martens2016mining} found that using validation data to train "for each lift threshold performs quite well for the corresponding lift threshold". It would be useful to study ways of optimizing classifiers for lift by directly modifying the classifier's optimized objective function, similar to the relationship found between RankBoost and AUC \citep[e.g.,][]{cortes2004auc,rudin2009margin}. 

A second direction is studying resource-constraint performance in the case of numerical outcomes. For example, in insurance, actuaries use numerical outcomes such as frequency and severity of claims. Questions of interest include how lift for numerical outcomes contrast with approaches such as Regression Receiver Operating Characteristic (RROC) \citep{hernandez2013roc}? how does lift relate to RMSE? Are there advantages of discretizing a numerical outcome into a binary one \citep[as in][]{rosset2007ranking}?

We note that recent increases in resource-constrained applications, aimed at efficiency and profitability, often carry the price of creating a larger social, economic, or human divide. This occurs, for example, when instead of prioritizing an intended action across records (e.g. prioritizing care in an emergency department), the action is applied only to a selected subset ("VIP treatment"). Such considerations are often reflected by laws and regulations that mandate offering a service to everyone, even if at different rates or conditions (e.g. insurance or legal protection). This work is not intended to change decision makers' goals from prioritization (ranking) to VIP treatment (resource constraining), but rather to point out the difference between the two and to better align the data mining performance measures accordingly.

We hope that this paper motivates researchers and practitioners using classifiers to consider whether their intended application is resource-constrained, and if so, to use gains and lift charts as the appropriate performance measures.

\section*{Acknowledgements}
I thank Sz-Wei (Sandy) Wu, Foster Provost, Boaz Shmueli, and Travis Greene for their valuable insights and feedback. Table 1, Figure 1, and the two example datasets were used as part of Sz-Wei Wu's Masters' thesis. Figure 2 is reproduced from \citet{provost2013data} with permission. 

\bibliographystyle{apa}  
\bibliography{references}  

\end{document}